\documentclass{article}
\usepackage{ijcai20}

\usepackage{times}

\usepackage{soul}
\usepackage{url}
\usepackage[hidelinks]{hyperref}
\usepackage[utf8]{inputenc}
\usepackage[small]{caption}
\usepackage{graphicx}
\usepackage{amsmath}
\usepackage{booktabs}
\usepackage{mathrsfs} 
\usepackage{amsfonts}
\usepackage[T1]{fontenc}
\usepackage{color,soul}

\usepackage{gensymb}
\usepackage{bbding}
\usepackage{pifont}

\newcommand{\norm}[1]{\left\lVert#1\right\rVert}
\newcommand{\cmark}{\ding{51}}%
\newcommand{\xmark}{\ding{55}}%

\usepackage{booktabs}
\usepackage{multirow}
\usepackage{lscape}
\usepackage{adjustbox}
\usepackage{tikz}
\usepackage{textpos}

\title{A Comparison of Uncertainty Estimation Approaches in Deep Learning Components for Autonomous Vehicle Applications}

\author{
	Fabio Arnez$^1$\and
	Huascar Espinoza$^1$\and
	Ansgar Radermacher$^1$\And
	François Terrier$^1$\\
	\affiliations
	$^1$CEA LIST, Gif-sur-Yvette, France\\
	\emails
	\{fabio.arnez, huascar.espinoza, ansgar.radermacher, francois.terrier\}@cea.fr
}

\newcommand\copyrighttext{%
	\footnotesize Copyright © 2020 for this paper by its authors. Use permitted under Creative Commons License Attribution 4.0 International (CC BY 4.0).}
\newcommand\copyrightnotice{%
	\begin{tikzpicture}[remember picture,overlay]
	\node[anchor=south,yshift=30pt] at (current page.south) {{\copyrighttext}};
	\end{tikzpicture}%
}

\begin{document}

\maketitle

\copyrightnotice


\begin{abstract}
A key factor for ensuring safety in Autonomous Vehicles (AVs) is to avoid any abnormal behaviors under undesirable and unpredicted circumstances. As AVs increasingly rely on Deep Neural Networks (DNNs) to perform safety-critical tasks, different methods for uncertainty quantification have recently been proposed to measure the inevitable source of errors in data and models. However, uncertainty quantification in DNNs is still a challenging task. These methods require a higher computational load, a higher memory footprint, and introduce extra latency, which can be prohibitive in safety-critical applications. In this paper, we provide a brief and comparative survey of methods for uncertainty quantification in DNNs along with existing metrics to evaluate uncertainty predictions. We are particularly interested in understanding the advantages and downsides of each method for specific AV tasks and types of uncertainty sources.
\end{abstract}


\section{Introduction}

In the last decade, Deep Neural Networks (DNNs) have witnessed great advances in real-world applications like Autonomous Vehicles (AVs) to perform complex tasks such as object detection and tracking or vehicle control. Despite substantial performance improvements introduced by DNNs, they still have significant safety shortcomings due to their complexity, opacity and lack of interpretability \cite{mcallister2017concrete}. In particular, DNNs are brittle to operational domain shift and even small data corruption or perturbations \cite{kuutti2020survey}. This impedes ensuring the reliability of the DNNs models, which is a precondition for safety-critical systems to ensure compliance with automotive industry safety standards and avoid jeopardizing human lives.


A concrete safety problem is to detect abnormal situations under uncertain environment conditions and DNN-specific unpredictability. These situations are difficult to analyze during system development phases, in a way that they can be properly mitigated at a real-time scale. Indeed, although a DNN model achieves great performance in a validation set from its operation environment, it is currently impossible to test and provide the same performance guarantees in all the possible environment configurations the system could encounter in the real world \cite{kuutti2020survey}. A common practice to overcome this problem is to use runtime monitoring of DNN components, so that safety can be ensured even if the component was not fully validated at design time \cite{hennebenchmarking,koopman2019autonomous}.  A central aspect to enable DNN monitoring is to provide a runtime treatment of uncertainties associated with DNN's predictions \cite{mcallister2017concrete,koopman2019autonomous}.


In this paper, we review common uncertainty estimation methods for DNNs and compare their performance and benefits for different AV tasks. These methods offer a potential solution for runtime DNN confidence prediction and detection of Out-of-Distribution (OOD) samples, since prediction probability scores in DNNs do not provide a true representation of uncertainty \cite{mohseni2019practical}. However, these methods still demand a high computational load, incorporate extra latency, and require a larger memory footprint. We compare these factors since they can represent a major impediment in safety-critical applications with tight time constraints and limited computation hardware. We also briefly focus on surveying uncertainty metrics that evaluate the performance of quantification methods, as another critical factor to ensure safety in AV systems.


The remainder of the paper is structured as follows. Section 2 describes the sources of uncertainty in deep learning for AVs. Section 3 presents a comparison of recent works in AV tasks that include uncertainty estimation methods for DNNs. It provides a brief review of common uncertainty estimation methods in deep learning as well as metrics for predictive uncertainty evaluation in classification and regression tasks. Section 4 discusses the open challenges and possible directions for future work.



\section{Background}

\subsection{Sources of Uncertainty in Deep Learning for Autonomous Vehicles}

Autonomous vehicles have to deal with  dynamic, non-stationary and highly unpredictable operational environments. Taking into account all the details from the operational environment at design time is an intractable task. Instead, the operational environment is constrained in a way that it considers only a subset of all possible situations that the system can encounter in operation. This process is known as  Operational Design Domain (ODD) adoption \cite{koopman2019many}, and safety requirements are built on the top of the ODD specification.

\smallskip

Given the constrained operational environment within system ODD, ensuring safety in an AV requires the identification of unfamiliar contexts by modeling AV's uncertainty \cite{mcallister2017concrete}. However,  there are many factors, not only related to the environment, that affect the system performance by introducing some degree of uncertainty. \cite{czarnecki2018towards} identify a set of factors that contribute to uncertainty in the perception function in an AV, and in this manner affect its performance. From this set, we take special attention to sensor properties, model uncertainty, situation and scenario coverage, and operational domain uncertainty factors. In the context of DNNs, the first two factors can be modeled by using uncertainty estimation methods, while the last two correspond to some degree of dataset shift (i.e. breaking the \textit{independent and identically distributed} assumption between training and testing data.) and Out-of-Distribution (OOD) samples \cite{quionero2009dataset,mohseni2019practical}.

\smallskip

Sensor properties like range, resolution, noise characteristics, and calibration can influence the amount of information in the samples delivered to a machine learning model during training or testing. In consequence, the effect of these properties are captured as noise and ambiguities inherent to the obtained samples. This type of noise in the data is known as \textit{Aleatoric} uncertainty,  and represents the incapability of completely sensing all the details of the environment \cite{kendall2017uncertainties,lee2019early,gustafsson2019evaluating}. Aleatoric uncertainty can be further further classified into \textit{homoscedastic} uncertainty (uncertainty that remains constant for different samples), and \textit{heteroscedastic} uncertainty (uncertainty that can vary between samples).

\smallskip

Model uncertainty is often referred to as \textit{Epistemic} uncertainty, and accounts for uncertainty in the model parameters. This type of uncertainty captures the ignorance of the model as a consequence of a dataset that does not represent the ODD well, or that is not sufficiently large \cite{kendall2017uncertainties,lee2019early}. Epistemic uncertainty is expected to increase in unknown situations (e.g. different environment ODD conditions such as weather or lightning), and can be explained away by incorporating more data.

\smallskip

Situation and scenario coverage is related to the degree in which situations and scenarios from an ODD are reflected in training and operation stages; while operational domain uncertainty refers to a discrepancy between ODD situations and scenarios present at training and those encountered at operation (e.g. scenarios from two different ODDs) \cite{czarnecki2018towards}. In both cases, uncertainty can be reduced by incorporating more data, or by adjusting the ODD specification.  However, it is extremely important to detect and discover OOD samples (i.e. outliers), especially those that have not been seen before, since those can lead to highly confident predictions that are wrong, i.e., the \textit{unknown-unknowns} \cite{bansal2018coverage}.

\medskip

In a similar fashion as the cases presented before, automotive industry standard ISO/PAS 21448 or SOTIF (Safety Of The Intended Functionality) \cite{iso2019pas}, provides a process to identify unknown and potentially unsafe scenarios to minimize the risk by recognizing the performance limitations from sensors, algorithms, or user misuse. Unsafe scenarios can be further classified into unsafe-known (e.g out of ODD samples) or unsafe-unknown (e.g. OOD samples).  Once an unknown-unsafe scenario or situation is identified, it becomes a known-unsafe scenario that can be mitigated at design time \cite{rauapproach,mohseni2019practical}.

\medskip

\smallskip
\subsection{Uncertainty Estimation Methods for DNNs}


In recent years, many probabilistic deep learning methods have been proposed to obtain an uncertainty measure from an approximation to the (highly multi-modal) predictive distribution, as well as methods for calibrating the outputs of DNNs. In general, there are two approaches for DNN predictive uncertainty calculation: sampling-based and sampling-free methods. Sampling-based methods rely on taking multiple predictive samples based on the same input to get the estimator that will be associated with uncertainty. Sampling-free methods require one single predictive output. These methods are further discussed in Section 3.

\medskip
\smallskip

\subsubsection{Neural Network Calibration}

Confidence calibration represents the degree to which a model’s predicted probability estimates the true correctness likelihood \cite{guo2017calibration}. Under ideal circumstances, we expect that the normalized outputs from a DNN (i.e softmax outputs) correspond to the true correctness likelihood \cite{guo2017calibration}. From a frequentist perspective, this can be viewed as a discrepancy measure between local confidence (or uncertainty) predictions and the expected performance in the long-run \cite{hubschneider2019calibrating,lakshminarayanan2017simple}. For example, we expect that a class predicted with probability $p$ is correct $p$\% of the time, i.e. from 100 samples predicted with confidence 0.9, we expect 90 correct predictions. DDNs can be calibrated by using \textit{Temperature Scaling}, a simple post-processing technique \cite{guo2017calibration}, or more recently, \textit{Dirichelt calibration} \cite{kull2019beyond}.  For a regression setting, \cite{kuleshov2018accurate,hubschneider2019calibrating} formalize the calibration notion for continuous variables, in which a $p$\% confidence interval should contain the true outcome $p$\% of the time.

\smallskip
\smallskip

Despite the improvements achieved with calibration methods, they can not be seen as a complete solution for uncertainty estimation problem, since calibration is performed relative to a validation dataset \cite{kull2019beyond,ashukha2020pitfalls} (i.e., calibration methods rely on in-distribution samples to learn a calibration map). In the presence of OOD samples, a model is no longer calibrated. This limits the contribution of calibration techniques to scenarios where huge training datasets are available.

\vfil


\section{Comparison of Uncertainty Estimation Methods in AV Domain}

In this section, we compare and analyze some common uncertainty estimation methods in terms of out-of-the-box calibration in the predictions (i.e. without a prior calibration), computational budget, memory footprint, and required changes in the DNN for applying each method (architecture, loss function, and others). We have chosen the most representative works to the best of our knowledge in each application. Some of the listed works introduce improvements by performing combinations between other methods. This is summarized in Table \ref{tab:UEComp}.

%
%

\subsection{Methods Limited to Aleatoric Uncertainty}

The first four methods listed in Table \ref{tab:UEComp} exclusively deal with aleatoric uncertainty. In classification tasks, uncertainty is usually represented by normalized logits at the output layer (e.g. softmax output) which can be interpreted as a probability distribution related to aleatoric uncertainty \cite{gustafsson2019evaluating}. Unfortunately, normalized outputs as probability distributions fail to capture model uncertainty and this very often results in overconfident predictions that are wrong \cite{guo2017calibration}, especially in the presence of dataset-shift. To overcome the problems of softmax, \cite{gast2018lightweight} propose to use a Dirichlet distribution instead. 

In a regression configuration, deep learning models do not have an uncertainty representation by default. The outputs of a DNN are intended to parameterize a probability distribution (e.g., Gaussian, Laplace) to obtain a probabilistic representation. This modification of the architecture allows DNNs to learn aleatoric uncertainty from the data itself by using thes heteroscedastic loss and maximum likelihood \cite{kendall2017uncertainties,ilg2018uncertainty}. Similarly, in the heteroscedastic version of the classification, \cite{kendall2017uncertainties} place a Gaussian distribution over the output logits (i.e., each logit with its respective variance), before the softmax layer is applied. An alternative approach replaces the input, output and activation functions of a DNN with probability distributions \cite{gast2018lightweight}. This method allows the propagation of a fixed uncertainty at the input to the output of the DNN employing \textit{Assumed Density Filtering} (ADF).

\subsection{Bayesian Neural Networks}
 Bayesian Neural Networks (BNNs), aim to learn a distribution over the weights instead of  point estimates. In this way, we look for the posterior distribution of the weights given the data $p(\mathbf{w}|\mathcal{D})$, by applying Bayes' theorem from the data likelihood and a chosen prior distribution over the weights $p(\mathbf{w})$:
\begin{equation}
p(\mathbf{w}|\mathcal{D}) = \frac{p(\mathcal{D}|\mathbf{w}) p(\mathbf{w})}{p(\mathcal{D})} = \frac{p(\mathcal{D}|\mathbf{w}) p(\mathbf{w})}{\int p(\mathcal{D}|\mathbf{w}) p(\mathbf{w}) d\mathbf{w}} \label{eq:weightPosterior}
\end{equation}
Given the predictive posterior distribution $p(\mathbf{w}|\mathcal{D})$, we obtain the  predictive posterior  distribution for a new input $\mathbf{x^{*}}$ by marginalizing over the model parameters:
\begin{equation}
p(\mathbf{y^{*}}|\mathbf{x^{*}},\mathcal{D} ) = \int p(\mathbf{y^{*}}|\mathbf{x^{*}},\mathbf{w}) p(\mathbf{w}|\mathcal{D})  d\mathbf{w} \label{eq:postPredDist}
\end{equation}
Instead of relying on only one configuration of the weights, we use every possible configuration of the weights (all possible models) weighted by the posterior on the parameters, to make a prediction, i.e. $p(\mathbf{y^{*}}|\mathbf{x^{*}}, \mathcal{D}) = \mathbb{E}_{p(\mathbf{w}|\mathcal{D})}[p(\mathbf{y^{*}}|\mathbf{x^{*}},\mathbf{w})]$. This represents the Bayesian Model Average (BMA) and accounts for \textit{epistemic uncertainty} \cite{wilson2020bayesian,gal2016uncertainty,blundell2015weight}.

Unfortunately, the integrals from \eqref{eq:weightPosterior} and \eqref{eq:postPredDist}  are intractable. Thus, we must build a distribution that approximates the true posterior distribution on the weights, $q(w) \approx p(\mathbf{w}|\mathcal{D}) $. Two main paradigms exist to build $q(w)$: \textit{Markov Chain Monte Carlo} (MCMC) and \textit{Variational Inference} (VI) methods. In the former, the gold standard is Hamiltonian Monte Carlo (HMC), and other methods like Stochastic Gradient MCMC (SG-MCMC) have been explored. However, MCMC methods are in general hard to scale to large DNNs due to the high-dimensional and multi-modal posterior distribution \cite{gustafsson2019evaluating}. In the latter case, VI methods approximate the posterior over the weights by approximating a simpler distribution $q_\phi(\mathbf{w})$ (e.g. a gaussian) parameterized by $\phi$. The parameters of $q_\phi(\mathbf{w})$ are found by minimizing the KL-divergence to $p(\mathbf{w}|\mathcal{D})$.

A particular scalable and easy to implement sample-based method for approximate VI is Monte Carlo Dropout (MCD) \cite{gal2016dropout}. In this method, dropout regularization is also applied at test time, so that $q_\phi(\mathbf{w})$ is a Bernoulli distribution. Dropout is only performed in some of  the deeper layers of  the DNN to model better high-level features and to avoid slow training \cite{mukhoti2018evaluating,kendall2015bayesian}. Dropout probabilities can be set manually, or the network can tune dropout rates during training \cite{gal2017concrete}.

All the MCD-related methods listed in Table \ref{tab:UEComp} refer to this approximation of BNNs. It can be noted from the performance comparison criteria, that the need to take multiple forward passes (output samples) for the same input to approximate the distribution from Equation \ref{eq:postPredDist} represents a major impediment to safety-critical applications with tight time constraints and limited computation hardware.

 
To get a representation of both types of uncertainty (aleatoric and epistemic), the methods presented in Section 3.1 have been used in combination with MCD. For example, in a regression configuration, a set of $T$ samples are taken from the predictions of a DNN that parameterize a distribution in its output: $\{\mathbf{\hat{y}_t},\hat{\sigma}_t\}_{t=1}^{T}$. However, since aleatoric uncertainty is learned from the data itself (by using the heteroscedastic loss), this approach could produce wrong uncertainty estimations in samples that include a higher level of uncertainty than that observed during training. Another approach presented in \cite{loquercio2020general},
applies MCD to take samples from a DNN where the input, output and activation functions are replaced by probability distributions according to \cite{gast2018lightweight}. This method permits uncertainty propagation at the input to the output of the DNN using ADF (e.g., sensor noise can be propagated to the output of the DNNs). This is an appealing method for AV applications where sensor properties are commonly known. Interestingly, the authors show that this method can be applied to trained DNNs and is architecture agnostic.


\begin{landscape}
	\begin{table}[h!]
		\begin{adjustbox}{max width=650pt, right}
			\begin{tabular}{@{}lp{0.45\textwidth}cccccc@{}}
				\toprule
				\multicolumn{1}{c}{\multirow{2}{*}{\textbf{Method}}} &
				\multicolumn{1}{c}{\multirow{2}{*}{\textbf{Autonomous Vehicle Task}}} &
				\multicolumn{2}{c}{\textbf{Uncertainty Captured}} &
				\multicolumn{4}{c}{\textbf{Comparison Criteria}} \\ \cmidrule(l){3-8} 
				\multicolumn{1}{c}{} &
				\multicolumn{1}{c}{} &
				\multicolumn{1}{c}{\textbf{Aleatoric}} &
				\multicolumn{1}{c}{\textbf{Epistemic}} &
				\textbf{\begin{tabular}[c]{@{}c@{}}Out-of-the box\\ Calibration\end{tabular}} &
				\textbf{\begin{tabular}[c]{@{}c@{}}Computational\\ Budget\end{tabular}} &
				\textbf{\begin{tabular}[c]{@{}c@{}}Memory\\ Footprint\end{tabular}} &
				\textbf{\begin{tabular}[c]{@{}c@{}}Changes in\\ DNN \end{tabular}} \\\specialrule{2pt}{0pt}{0pt}
				\textbf{\begin{tabular}[c]{@{}l@{}}Softmax logits \\ as parameters \\ of a prob. dist.\end{tabular}} &
				- Object Detection \cite{feng2019leveraging} &
				\cmark&
				\xmark &
				Bad &
				Fair &
				Slow &
				Small \\ \cmidrule(l){2-8} 
				\textbf{\begin{tabular}[c]{@{}l@{}}Outputs as \\ parameters\\ of a prob. dist.\end{tabular}} &
				- Object Detection \cite{feng2019leveraging} &
				\cmark&
				\xmark &
				Bad &
				Fair &
				Slow &
				Small \\ \cmidrule(l){2-8} 
				\textbf{\begin{tabular}[c]{@{}l@{}}Inputs,  activation \\ and output as\\ prob. dist. \& ADF\end{tabular}} &
				- Optical Flow \cite{gast2018lightweight} &
				\cmark&
				\xmark &
				Undefined &
				Low &
				Low &
				Mid \\ \cmidrule(l){2-8} 
				\textbf{\begin{tabular}[c]{@{}l@{}}Point estimate \\ \& MCD\\ regression\end{tabular}} &
				- Steering Angle Prediction \cite{hubschneider2019calibrating,michelmore2018evaluating,michelmore2019uncertainty} &
				\xmark&
				\cmark &
				Fair &
				Fair &
				Low &
				None \\ \cmidrule(l){2-8} 
				\textbf{Softmax  \& MCD} &
				\begin{tabular}[c]{@{}p{0.45\textwidth}@{}}- Traffic Sign Recognition \cite{hennebenchmarking}\\ - Semantic Segmentation \cite{phan2019bayesian,mukhoti2018evaluating,gustafsson2019evaluating}\end{tabular} &
				\xmark &
				\cmark&
				Fair &
				Fair &
				Low &
				None \\ \cmidrule(l){2-8} 
				\textbf{\begin{tabular}[c]{@{}l@{}}Deep \\ Ensembles\end{tabular}} &
				\begin{tabular}[c]{@{}p{0.45\textwidth}@{}}- Steering Angle Prediction \cite{hubschneider2019calibrating}\\ - Traffic Sign Recognition \cite{hennebenchmarking}\\ - Semantic Segmentation \cite{gustafsson2019evaluating}\\ - Depth Estimation \cite{gustafsson2019evaluating}\end{tabular} &
				\cmark&
				\cmark&
				Good &
				High &
				High &
				Small \\ \cmidrule(l){2-8} 
				\textbf{\begin{tabular}[c]{@{}l@{}}Bootstrap \\ Ensembles\end{tabular}} &
				\begin{tabular}[c]{@{}p{0.45\textwidth}@{}}- Steering Angle Prediction \cite{hubschneider2019calibrating}\\ - Optical Flow \cite{ilg2018uncertainty}\end{tabular} &
				\cmark&
				\cmark&
				Bad &
				Fair &
				Fair &
				Mid \\ \cmidrule(l){2-8} 
				\textbf{\begin{tabular}[c]{@{}l@{}}Softmax logits \\ as parameters \\ of a prob. dist. \\ \& MCD\end{tabular}} &
				- Object Detection \cite{feng2018towards} &
				\cmark&
				\cmark&
				Fair &
				High &
				Low &
				Small \\ \cmidrule(l){2-8} 
				\textbf{\begin{tabular}[c]{@{}l@{}}Outputs as \\ parameters of \\ a prob. dist. \\ \& MCD\end{tabular}} &
				\begin{tabular}[c]{@{}p{0.45\textwidth}@{}}- Object Detection \cite{feng2018towards}\\ - Steering Angle Prediction \cite{lee2019early,lee2019ensemble,lee2019perceptual}\\ - Depth Estimation \cite{gustafsson2019evaluating}\end{tabular} &
				\cmark&
				\cmark&
				Fair &
				High &
				Low &
				Small \\ \cmidrule(l){2-8} 
				\textbf{\begin{tabular}[c]{@{}l@{}}Inputs,  activation \\ and output as\\ prob. dist. \& ADF \\ \&MCD\end{tabular}} &
				- Steering Angle Prediction \cite{loquercio2020general} &
				\cmark&
				\cmark&
				Undefined &
				High &
				Low &
				Mid \\ \cmidrule(l){2-8} 
				\textbf{MDNs} &
				\begin{tabular}[c]{@{}p{0.45\textwidth}@{}}- Steering Angle Prediction \cite{hubschneider2019calibrating,choi2018uncertainty}\\ - Future Prediction \cite{makansi2019overcoming}\end{tabular} &
				\cmark&
				\cmark&
				Bad &
				Low &
				Low &
				None \\ \cmidrule(l){2-8} 
				\textbf{\begin{tabular}[c]{@{}l@{}}MDNs \\ with stages\end{tabular}} &
				- Future Prediction \cite{makansi2019overcoming} &
				\cmark&
				\cmark&
				Undefined &
				Low &
				Low &
				High \\\specialrule{1pt}{0pt}{0pt}
			\end{tabular}
		\end{adjustbox}
		\caption{Uncertainty Estimation Methods Comparison}
		\label{tab:UEComp}
	\end{table}
	
\end{landscape}

\subsection{Deep Ensembles}
A Deep Ensemble (DE) is another sample-based method, in which $M$ DNNs are trained to obtain the predictive distribution $p(\mathbf{y|x})$ \cite{lakshminarayanan2017simple}. Each DNN learns a set of parameters $\mathbf{w}$ that are point estimates, starting for different random initialization and repeating the minimization $M$ times. In an ensemble, predictions are averaged and can be considered as a mixture model that is equally weighted:
\begin{equation}
p(\mathbf{y|x}) = \frac{1}{M} \sum_{i=1}^{M} p(\mathbf{y|x, \hat{w}}^i), \; \{\mathbf{\hat{w}}^{(i)}\}_{i=1}^{M}\label{eq:de}
\end{equation}

For classification, equation \eqref{eq:de} corresponds to an average of the softmax probabilities. For regression, the outputs that parameterize a probability distribution are averaged to represent the mean and variance of the mixture. In this manner, both types of uncertainty (aleatoric and epistemic) can be easily captured. Although DE is considered a non-Bayesian method, expression \eqref{eq:de} represents an approximation of \eqref{eq:postPredDist} since $\{\mathbf{\hat{w}}^{(i)}\}_{i=1}^{M}$ can be seen as samples taken from distribution that approximates the true posterior, by exploring different modes of from $p(\mathbf{w}|\mathcal{D})$ \cite{fort2019deep,wilson2020bayesian}.

As presented in Table \ref{tab:UEComp}, the DE method tends to outperform approximate Bayesian inference methods like MCD, for both, uncertainty estimates and accuracy \cite{gustafsson2019evaluating}. A recent work from \cite{snoek2019can} also shows, that DE is more robust to dataset shift. These works suggest that DE  should be considered as the new standard method for predictive distributions and uncertainty estimation. However, DE has some drawbacks, especially if the target application is a safety-critical application. DE requires a higher computational load and a larger memory footprint, as shown in Table \ref{tab:UEComp}. For the training and testing stage, the number of parameters, and the inference times scale linearly with $M$. To mitigate this problem, \cite{osband2016deep} propose a fused version of ensembles with multiple heads. All the heads share the convolutional layers (feature extractors) and each head is trained using boostrap samples.

\subsection{Mixture Density Networks}
Mixture Density Networks (MDN) \cite{bishop1994}, is a sample-free method for regression tasks, where the aim is to train a DNN that predicts the parameters of a Gaussian Mixture Model (GMM)  given an input $\mathbf{x}$. A GMM is formed by a weighted sum of $K$ Gaussians, to model the conditional distribution:
\begin{equation}
p(\mathbf{y|x})= \sum_{i=1}^{K}\pi_{i}(\mathbf{x})\mathcal{N}(\mathbf{y}|\mu_{i}(\mathbf{x}),\Sigma_{i}(\mathbf{x}))\label{eq:gmm}
\end{equation}
where $\pi_{i}(\mathbf{x}),\mu_{i}(\mathbf{x}),\Sigma_{i}(\mathbf{x})$ represent the set of parameters of the GMM as a function of the input $\mathbf{x}$ for $K$ mixtures. For training, \textit{Negative Log-likelihood} (NLL) is used as loss function.

By using the law of total variance, \cite{choi2018uncertainty} formalized the acquisition of aleatoric and epistemic uncertianty in MDNs. As a first step, the expectation of the GMM is obtained as a combination of the mixture components in a weighted sum: $\mathbb{E}[\mathbf{y|x}] = \sum_{i=1}^{K}\pi_{i}(\mathbf{x}) \mu_{i}(\mathbf{x})$. The predicted variance is composed of the weighted sum of the variances and the weighted variances of the means:
\begin{equation}
\resizebox{.91\linewidth}{!}{$
	\displaystyle
\mathbb{V}[\mathbf{y|x}] = \sum_{i=1}^{K}\pi_{i}(\mathbf{x})\Sigma_{i}(\mathbf{x}) + \sum_{i}^{K}\pi_{i}(\mathbf{x})  \norm{ \mu_{i}(\mathbf{x})  - \sum_{i}^{K} \pi_{i}(\mathbf{x}) \mu_{i}(\mathbf{x}) }^2
$}
\end{equation}
where the first term represents the aleatoric uncertainty and the second term represents the epistemic uncertainty. We refer the reader to \cite{choi2018uncertainty} for more details about uncertainty acquisition in MDNs.

As pointed out in Table \ref{tab:UEComp}, the sampling-free nature of this method reduces the computation load, memory footprint, and permits complex distribution modeling with respect to the methods described before. These characteristics are attractive for real-time applications. However, MDNs suffer from numerical instability for high dimensional problems and mode collapse when using regularization techniques \cite{makansi2019overcoming}.


\subsection{Quality Metrics for Uncertainty Estimation} 
In this section, we discuss common metrics for evaluating the quality of uncertainty estimation.

\paragraph{Classification Metrics.} Different methods for uncertainty estimation exist for classification tasks. \textit{Variation Ratio} and information metrics such as \textit{Predictive Entropy, Mutual Information}, can be used in classification settings to represent uncertainty \cite{gal2016uncertainty}. Variation ratio is a measure of dispersion; mutual information captures model confidence, and predictive entropy accounts for epistemic and aleatoric uncertainty \cite{mukhoti2018evaluating,michelmore2018evaluating,phan2019bayesian}. \cite{mukhoti2018evaluating} propose specific performance metrics for semantic segmentation to evaluate Bayesian models. Since there is no ground-truth for uncertainty estimation, \cite{snoek2019can,lakshminarayanan2017simple} argue that proper scoring rules are NLL and \textit{Brier} score. NLL depends on predictive uncertainty and is commonly evaluated in a held-out set, however, it can overestimate tail probabilities; whereas Brier-score measures the accuracy of predictive probabilities by a sum of squared differences between the predicted probability vector and the target, nonetheless, this score is prone to avoid capturing infrequent events. Other evaluation metrics independent of score values are: the \textit{Area Under the Receiver Operating Characteristic} (AUROC), \textit{Area Under Precision Recall Curve} (AUPRC), and Area Under Risk-Coverage (AURC) \cite{hendrycks2016baseline,ding2019evaluation}.


\paragraph{Regression Metrics.} Similarly, in regression tasks, NLL is a proper scoring rule for a likelihood that follows Gaussian distribution \cite{lakshminarayanan2017simple,kendall2017uncertainties}. Furthermore, \cite{ilg2018uncertainty} introduces a relative measure for uncertainty estimation, the \textit{Area Under the Sparsification Error} (AUSE) curve, that measures the difference between the dispersion of predictions (affected by predictive uncertainty), and a oracle in terms of true prediction error, e.g. \textit{Root Mean Squared Error} (RMSE) \cite{gustafsson2019evaluating}.

\paragraph{Calibration Metrics.} For classification tasks, common quality metrics are \textit{Expected Calibration Error} (ECE) and \textit{Maximum Calibration Error} (MCE) \cite{guo2017calibration}. The former measures the difference between expected accuracy and expected confidence; the latter identifies the largest discrepancy between accuracy and confidence, which is of particular interest in safety-critical applications. For a regression configuration, \cite{kuleshov2018accurate} use calibration error as a metric that represents the sum of weighted squared differences between the expected and observed (empirical) confidence levels; correspondingly in \cite{gustafsson2019evaluating}, the authors propose to use the \textit{Area Under the Calibration Error} curve (AUCE) as an absolute measure of uncertainty. The before-mentioned authors use reliability diagrams (i.e. calibration plots) to get a visual representation of model calibration. Regardless of drawbacks with OOD samples, calibration plots and measures are used extensively to compare the predictive quality of other uncertainty estimation methods.




\subsection{Considerations per AV Task Type}



In the context of AVs, for (end-to-end) steering angle prediction, a broad variety of uncertainty estimation methods have been applied. In some works only epistemic uncertainty was captured by using MCD \cite{michelmore2018evaluating,michelmore2019uncertainty}. However, usually both types of uncertainty are captured \cite{lee2019early,lee2019ensemble,lee2019perceptual} by using the method proposed by  \cite{kendall2017uncertainties}, or by using DE, boostrap ensembles, or MDNs. The calibration plots presented in \cite{hubschneider2019calibrating} show that MCD has better out-of-the-box calibration than bootstrap ensembles or MDNs; the last two methods are overconfident in their predictions. In this particular task, safety mechanisms have been proposed when uncertainty estimations surpass a given or learned threshold in order to improve vehicle safety \cite{michelmore2018evaluating,michelmore2019uncertainty,lee2019early}.

Under the modular pipeline paradigm for AV control, probabilistic modeling has mainly been applied to perception tasks like object detection from 3D Lidar, semantic segmentation and depth estimation. For 3D object detection from Lidar point-clouds, \cite{feng2018towards} estimate aleatoric and epistemic uncertainty  using the methods proposed by \cite{kendall2017uncertainties}. However, epistemic uncertainty estimation with MCD introduces a high computational cost. A later work from  \cite{feng2019leveraging} leverages aleatoric uncertainties to greatly improve the performance and reduce the computational load from MCD. In \cite{feng2019can} the authors show that predictions for classification and regression are  miscalibrated, and propose methods to fix calibration of DNNs and produce better uncertainty estimates.

For semantic segmentation, \cite{phan2019bayesian,mukhoti2018evaluating,gustafsson2019evaluating} model aleatoric uncertainty from the softmax output, and epistemic uncertainty by using MCD or ensembles.  Common uncertainty metrics in this case are predictive entropy and mutual information \cite{mukhoti2018evaluating}. For Depth estimation, \cite{gustafsson2019evaluating} compares DE with the heteroscedastic regression in combination with MCD \cite{kendall2017uncertainties}. In both previous tasks (semantic segmentation and depth estimation) DE achieves better performance and calibration than MCD variants \cite{gustafsson2019evaluating}. However, in DE the computational cost at training and testing grows linearly with the number of ensembles. Similarly for traffic sign recognition, DE exhibit the best-calibrated outputs, but in this case, MCD in combination with softmax also produces well-calibrated outputs close to that from DE \cite{hennebenchmarking}.

For optical flow, \cite{gast2018lightweight} capture aleatoric uncertainty by replacing the input, output and activation functions with probability distributions. This method allows propagating a fixed value of uncertainty at the input to the output of the DNN. \cite{ilg2018uncertainty} present an alternative approach, where DE and bootstrap ensembles were used to obtain the predictive uncertainty.

For future  prediction, \cite{makansi2019overcoming} propose an improvement to MDNs to predict the multi-modal distribution of positions of a vehicle in the future. This method presents two stages: a sampling and a fitting network. The former network receives the current position of the vehicle as an input and
outputs a fixed number of hypotheses for future positions. The latter network fits a mixture distribution to the hypothesis estimated in the first network. This improvement helps to avoid mode collapse in MDNs, however, high dimensional outputs remain challenging for this approach.



\section{Conclusions}

We presented a comparative survey for uncertainty estimation methods for both, classification and regression tasks, in the AV domain. We also provide a general comparative analysis of these methods. From this analysis we can see that DE  has become a gold-standard for uncertainty quantification in many AV tasks thanks to its high-quality uncertainty predictions and its robustness to OOD samples. However, the high computational load and large memory footprint, can hinder its use in safety-critical applications that have hardware limitations or tight time-constraints. Here,  sampling-free methods are an interesting avenue for future research. New robust (to OOD) and lightweight approaches should be explored in the AV domain, to produce good-quality uncertainty estimates. We also observed that predictions from these methods are uncalibrated (overconfident or underconfident) and are usually applied to classification tasks. We encourage the application of calibration methods also for regression tasks by using the methods proposed by \cite{kuleshov2018accurate} instead of limiting the assessment of predictions with only reliability diagrams. We also suggest to study and compare uncertainty estimation methods under dataset-shift conditions to assess their robustness. For future work, we plan to incorporate uncertainty information into the \textit{Responsability-Sensitive Safety} model \cite{shalev2017formal}. This generalizes the approach from \cite{salay2020purss} by considering component uncertainty from different AV subsystems and propagating it through them. These subsystems could include DNNs e.g. for planning and control.


%


\section*{Acknowledgments}

This work has received funding from the COMP4DRONES project, under Joint Undertaking
(JU) grant agreement N\degree 826610. The JU receives support from the European Union’s Horizon 2020 research and innovation programme and from Spain, Austria, Belgium, Czech Republic, France, Italy, Latvia, Netherlands.

\bibliographystyle{named}
\bibliography{ijcai20}
\end{document}